\documentclass[letterpaper, 10 pt, conference]{ieeeconf}  

\IEEEoverridecommandlockouts                              

\usepackage{amssymb}  

\usepackage{subcaption}
\usepackage{color}
\usepackage{graphicx}
\usepackage{graphbox}
\usepackage{fixmath}
\usepackage{amsmath}

\usepackage{cleveref}
\usepackage{ifthen}
\usepackage{xspace}

\title{\LARGE \bf
A Study on Multirobot Quantile Estimation in Natural Environments
}

\author{Isabel M. Rayas~Fern\'andez$^{1}$, Christopher E. Denniston$^{1,\dagger}$, Gaurav S. Sukhatme$^{1,*}$
\thanks{*This work was supported in part by the Southern California Coastal Water Research Project Authority under prime funding from the California State Water Resources Control Board on agreement number 19-003-150 and in part by USDA/NIFA award 2017-67007-26154.}
\thanks{This material is based upon work supported by the National Science Foundation Graduate Research Fellowship Program under Grant No. DGE-1842487. Any opinions, findings, and conclusions or recommendations expressed in this material are those of the author(s) and do not necessarily reflect the views of the National Science Foundation.}
\thanks{$^{1}$Department of Computer Science, University of Southern California.}%
\thanks{$^{*}$G.S. Sukhatme holds concurrent appointments as a Professor at USC and as an Amazon Scholar. This paper describes work performed at USC and is not associated with Amazon.}%
\thanks{$^{\dagger}$C.E. Denniston is now at OffWorld, Inc. This paper describes work performed at USC and is not associated with OffWorld.}%
}

\newcommand{\gtlocations}{\mathbold{G^\#}}
\newcommand{\gtsensedlocations}{\mathbold{X^\#}}
\newcommand{\gtsensedvalues}{\mathbold{Y^\#}}
\newcommand{\gtlocation}{\mathbold{g^\#}}

\newcommand{\sensedlocations}{\mathbold{X}}
\newcommand{\sensedvalues}{\mathbold{Y}}
\newcommand{\location}{g}
\newcommand{\sensedlocation}{x}

\newcommand{\nrobots}{N}
\newcommand{\spread}{\alpha}

\newcommand{\quantiles}{Q}

\newcommand{\estimatedquantilevalues}{\tilde{V}}

\newcommand{\estimatedquantilevaluesfinal}{\estimatedquantilevalues_{\textrm{final}}}
\newcommand{\quantilevalues}{V}

\newcommand{\objectivefunction}{f}
\newcommand{\argmax}{\arg\!\max}

\newcommand{\numtiles}{|\quantiles|}

\newcommand{\boxheight}{0.45\linewidth}
\newcommand{\boxwidth}{0.49\linewidth}

\newcommand{\psig}{p \leq 5\textrm{e}{-2}} 

\newcommand{\pppsig}{p \leq 1\textrm{e}{-3}}

\newcommand{\wpres}[2]{T {=} #1,~p {\leq} #2}

\newif\ifshowrev
\newif\ifshowrevv

\newcommand*{\rev}[1]{
\ifshowrev
\leavevmode\unskip {\color{red}\textrm{#1}} \unskip
\else 
\leavevmode\unskip #1 \unskip
\fi 
}

\begin{document}

\maketitle
\thispagestyle{empty}
\pagestyle{empty}

\begin{abstract}
Quantiles of a natural phenomena can provide scientists with an important understanding of 
different
spreads of concentrations.
When there are several available robots,
it may be advantageous to pool resources in a collaborative way to improve performance. 
A multirobot team can be difficult to practically bring together and coordinate. 
To this end, we present a study across several axes of the impact of using multiple robots to estimate quantiles of a distribution of interest using an informative path planning formulation. We measure quantile estimation accuracy with increasing team size to understand what benefits result from a multirobot approach in a drone exploration task of analyzing the algae concentration in lakes. 
We additionally perform an analysis on several parameters, including the spread of robot initial positions, the planning budget, and inter-robot communication, and
find that while using more robots generally results in lower estimation error, this benefit is achieved under certain conditions. 
We present our findings in the context of real field robotic applications and discuss the implications of the results and interesting directions for future work.
\end{abstract}

\section{Introduction}


Scientists who study natural environments have used robots to assist in surveying or exploring regions of interest, for example to monitor harmful algal blooms \cite{kemna_pilot_2018,harmful2022}. To describe such phenomena both flexibly and in an interpretable manner, it has been proposed to specify quantiles of interest that robots can target during exploration \cite{rayas2022informative}. 
\textit{Quantile estimation} refers to acquiring the value of a given quantile in a distribution. For example, the median algae concentration would be given by the value of the 0.5 quantile. 
Previous work on quantile estimation in this context has focused on single-robot adaptive surveys; in this work, we study multi-robot surveys, motivated by groups that have more than one robot available, and by collaborative surveys between groups that pool their robot resources to maximize scientific output from a survey. 
Though the naive assumption may be that more robots will always be better, in this work, we aim to investigate, in a principled manner,  under what conditions this is true.
Deploying a robot in the environment can be expensive,
and it is unclear how having more robots in such a use case will scale.
This study assesses the impact of team size, starting location, planning budget, and communication on quantile estimation tasks in field environments. We believe it is an important first step toward principled decisions for field robot deployments in aquatic biology.

As our contributions, we present:
\begin{itemize}
    \item the first study on multirobot quantile estimation;
    \item quantitative results on the effect of team size on performance on real-world aquatic datasets;
    \item quantitative results on the effect of parameters including initial location spread, exploration budget, and inter-robot communication on performance, giving insight into what matters for a multirobot study;
    \item the results in the context of field applications and how they may impact different experimental setups.
\end{itemize}


\begin{figure}
    \centering
    \includegraphics[width=\columnwidth]{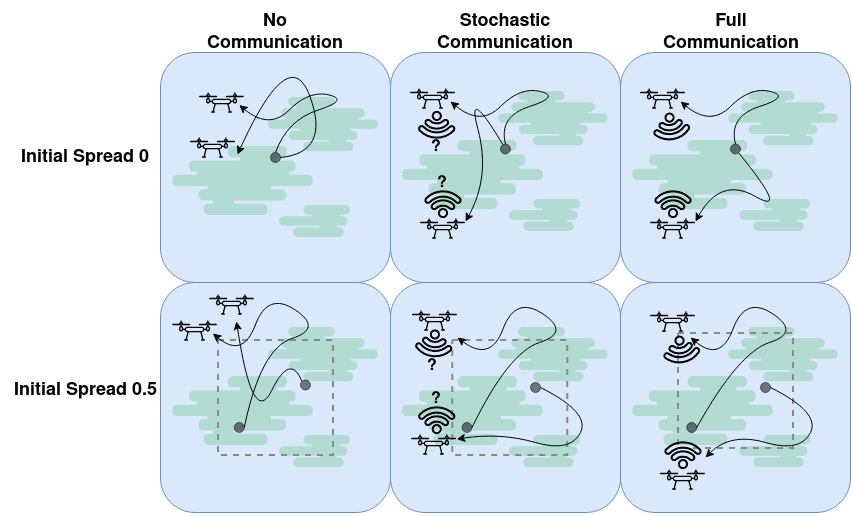}
    \caption{Some variations of multirobot planning approaches.
    Top row shows initial spread $\spread = 0$: All robots start in the center.
    Bottom row shows initial spread $\spread = 0.5$: Robots start spread in an area 50\% length and width of entire workspace.
    Left column shows no communication: Robots have no knowledge of the others; equivalent to each robot planning as if it were the only one.
    Middle column shows stochastic communication: Robots attempt to share observations but messages fail stochastically based on inter-robot distance.
    Right column shows full communication: Robots share the same environment model; equivalent to centralized planning.
    }
    \vspace{-10pt}
    \label{fig:hero}
\end{figure}

\section{Background}
We present relevant background on the motivating problem and place our contributions in the context of prior work.

\subsection{Environmental Quantile Estimation}
Monitoring and understanding algae growth in natural environments is the primary motivation for this work.
Spatial heterogeneity in the distribution of algal and cyanobacterial blooms in
freshwater and marine ecosystems is well known \cite{Seegers2015}.
Accurately characterizing variations is important to investigate average trends as well as dangerous conditions in our water systems, as exposure to cyanobacterial or algal toxins can be devastating to humans as well as animals \cite{Mehinto2021}.
Previous works have used robotic surveys as an integral part of studying and analyzing these environments and their algal and cyanobacterial characteristics \cite{Zhang2020, Sharp2021}, and recent work has specifically used quantile estimation to provide a flexible framework for these studies \cite{rayas2022informative}.

Here, we build upon the work in \cite{rayas2022informative}, but there are several differences.
First, \cite{rayas2022informative} seeks to produce, as a final output, a set of locations at which the desired quantile values can be found, while our goal is to accurately estimate the desired quantile values.
Second, we generalize the problem to the multirobot domain, where \cite{rayas2022informative} only considers one robot.
Finally, where \cite{rayas2022informative} proposes new objective functions for planning tailored to the quantile estimation problem, our focus is to investigate the impact of different factors on quantile estimation accuracy, and thus we adopt just one of the previously proposed objective functions in our work.

\subsection{Informative Path Planning (IPP)}
\label{ssec:ipp}
Informative path planning (IPP) is a common planning framework often used in an online manner which uses knowledge of the environment in the form of an internal model to inform the next action that is taken, after which new information is gained and incorporated into the model, and the process repeats.
IPP uses objective functions, which depend on the application, to determine which actions are the most informative at each step.
The robot produces a planned trajectory $p$ at each planning step which maximizes (or minimizes) the objective function $\objectivefunction$, and these trajectories together create the complete path $P$. 
Typically there is a planning step limit, or budget, $B$ on the problem which defines the maximum cost $c$ of the path \cite{denniston_icra_2020}. 
We describe this as
$P^* = \argmax_{P \in \Phi} f(P) ~ | ~  c(P) \leq B    $
where $\Phi$ is the space of full trajectories and $P^*$ is the optimal trajectory. 
One widely used internal model of the environment is a Gaussian Process (GP), which can represent a belief distribution over the environment and incorporates spatial dependency of the distribution into its predictions of unmeasured locations.
The GP produces an estimate of the mean value $\mu(\sensedlocation)$ and variance $\sigma^2(\sensedlocation)$ at a specific location $\sensedlocation$.
We pose our problem as an IPP problem, where the overall goal is to minimize error in the estimate of a set of quantiles of a distribution in the environment, using a GP to model the environment.

Partially Observable Markov Decision Processes (POMDPs) 
are frequently used to solve the planning problem \cite{borenstein_bayesian_2014, Marchanta}. 
POMDPs can represent uncertainty in the environment and simultaneously provide optimal actions given noisy observations, resulting in widespread applicability in planning problems. 
For these reasons, here we use the POMDP formulation to solve our IPP problem.

\subsection{Multirobot Studies}
\label{ssec:multirobot}
Multirobot systems research has a rich history; many well-known problems have been extended to more than one robot, such as multirobot SLAM \cite{howard2006multi}, multirobot exploration \cite{howard2006experiments}, and multirobot learning \cite{sartoretti2019distributed}.
In this work, we are interested in the problem of multirobot quantile estimation.
Specifically, we ask whether and under what conditions a multirobot approach to this problem is effective. 

Many previous studies concerning the effect of robot team size on performance have focused on human factors such as mental demand and operator workload \cite{wang2009search, velagapudi2008scaling},
while those that have studied autonomous group performance have shown inconclusive results regarding correlation between performance and team size \cite{rosenfeld2008study}, and that multirobot problems tend to follow the Law of Marginal Returns \cite{rosenfeld2004study, stachniss2008efficient}, which states that as more resources are added to a problem, smaller returns are generated. 
This property, also known as submodularity, can be seen both in number of robots as well as measurements taken, and previous work has exploited it to approximately solve planning problems \cite{Corah-2019-120007}.
Here, we seek to investigate the effect of using multiple homogeneous robots specifically on the task of quantile estimation in natural environments, with the aim of understanding how to most effectively use resources for challenging and resource-constrained field work problems.

\subsubsection{Multirobot IPP}
Previous works have used objective functions like entropy or mutual information to explore phenomena such as temperature fields, plankton density, salinity, and chlorophyll \cite{kemna_pilot_2018,guestrin_near-optimal_2005}.
Spatial correlation in the concentration has been exploited to solve the planning problem while improving the tradeoff between performance and efficiency \cite{cao2013multi}.
Planning with continuous connectivity constraints has also been explored.
In \cite{dutta2019multi}, bipartite graphs are used to determine where robots next visit. Though the main focus there is on varying the communication radius and the amount of information used to make decisions with, the authors also find a linear relationship between the number of robots and the improvement using an entropy-based objective function.
Bipartite graph matching has also been used to iteratively plan paths for robots to the most informative points in the environment \cite{ma2018multi}. 
A similar problem is addressed in \cite{dutta2020multi} where there is intermittent communication in cluttered environments. There, the region is partitioned into Voronoi cells to better balance the workload between robots when they come into communication range with each other. 
Repeated Voronoi partitioning has been used combined with limited information-sharing between underwater robots \cite{kemna2017multi}.
Such non-constant connectivity has also been studied in \cite{hollinger2010multi} where multirobot search with periodic connectivity is resolved using implicit coordination to address scalability.
Reinforcement learning has been used to learn a planning policy prioritizing exploring hotspots and robust to robot failures \cite{pan2022marlas}. 
Previous work in swarm robotics has investigated robot adaptability in tracking sites with time-varying quality \cite{talamali2021less} and the effect of swarm size on target tracking \cite{kwa2023effect}.

Our work differs from these in that it explicitly addresses the question of different team sizes and other factors' impact on performance, and is motivated by targeted field robotics deployments rather than the collective behavior exhibited by large swarms. Further, the task we study is to improve the estimates of an arbitrary set of quantiles, rather than coverage or hotspots which has typically been the focus of other work.


\section{Methods}
The problem considered in this paper is that of accurately estimating a set of quantiles of a distribution of interest, such as algae concentration in a lake, 
by exploring a finite 2D workspace with $\nrobots$ robots that take measurements at the locations they visit.
Formally, we discretize the planning space to $\gtlocations \subset
\mathbb{R}^2$, the set of locations that robots can visit defined by a grid.
Robots can move in the $\pm$ $x$ or $y$
directions.
There are $\nrobots$ collaborative homogeneous robots, each a fixed-height drone
which takes measurements via a downward pointing
hyperspectral camera.  Robots can take a picture at each location with
measurement resolution finer than $\gtlocations$, so $\gtsensedlocations$
represents the set of locations that a robot can measure by taking a picture at
each $\gtlocation \in \gtlocations$, and $\gtsensedvalues$ represents the actual
values at those locations.
The quantiles of interest $\quantiles$ are assumed to be given from the scientists or operators, and their true values are
defined as $\quantilevalues = \textit{quantiles}(\gtsensedvalues,\quantiles)$.
The planning step budget is $B_T$. The locations at which a robot has taken a
measurement up to time $t$ and the corresponding values measured are represented
by $\sensedlocations_{0:t}$ and $\sensedvalues_{0:t}$, respectively.  Each robot
$n$ maintains an estimate of the quantile values which is given by
    $\estimatedquantilevalues_n = quantiles(\mu^{(n)}_{GP_t}(\gtsensedlocations),\quantiles)$
where $\mu^{(n)}_{GP_t}(\gtsensedlocations)$ is the estimate of all possible locations
using the GP conditioned on $\sensedlocations_{0:t}$ and $\sensedvalues_{0:t}$. 
Note that, for clarity, we will omit the $(n)$ superscript in the remainder of the section.

For a given $\nrobots$ and $B_T$, our problem is to select a complete path
$\boldsymbol{P}^*$ composed of paths $P_n$ for each robot $n$ within a
budget constraint: 
%
$\boldsymbol{P}^* = \argmax_{\boldsymbol{P}} f(\boldsymbol{P})$
%
where
   $\boldsymbol{P} = \bigcup_{n \in \nrobots} P_n$,
%
   $ \forall n \in N :~\textit{length}(P_n) \leq B_n$,
and $f$ is the objective function used during planning. 
In this work, we aim to estimate the value of the underlying
concentration at $Q$. Note that our goal is not finding the highest
value nor optimizing model accuracy at every location.
Thus, we use the quantile standard error objective function to evaluate a
proposed location \cite{rayas2022informative}, which measures the difference in
standard error of the estimated quantile values using a robot's current environment
model $\mu_{GP_{i-1}}$, compared to using its current model updated with the
expected new measurements at the proposed location $\mu_{GP_{i}}$.
The second term encourages exploration of locations with
high model uncertainty weighted by the parameter $c$:
%
$
    \objectivefunction(\sensedlocations_i) =
\frac{d}{\numtiles} +
\sum_{\sensedlocation_j \in \sensedlocations_i} c\sigma^2(\sensedlocation_j)$
%
where
%
    $d =
\|se(\mu_{GP_{i-1}}(\gtsensedlocations),\quantiles) -
se(\mu_{GP_{i}}(\gtsensedlocations),\quantiles) \|_{1}$.

\begin{table}[]
\caption{Informative Path Planning as a POMDP \cite{rayas2022informative}.}
\begin{tabular}{c|c}
\textbf{POMDP} & \textbf{Informative Path Planning}        \\ \hline
States         & Robot position $\location_t$, Underlying unknown function $GT$ \\ \hline
Actions        & Neighboring search points                     \\ \hline
Observations   & Robot position $\location_t$, Measured location(s) \rev{$\sensedlocations_t = o(\location_t)$}   \\ 
               & Measured value(s) \rev{$\sensedvalues_t = GT(\sensedlocations_t) + \mathcal{N}(0,\sigma_\textrm{noise})$} \\\hline
Belief         & \rev{GP conditioned on previously measured }     \\ 
& \rev{locations  $(\sensedlocations_{0:t-1})$ and values $(\sensedvalues_{0:t-1})$} \\ \hline
Rewards        & \rev{$\objectivefunction(\sensedlocations_t)$}      
\end{tabular}
\label{tab:BayesianSearchAndPOMDPs}
\vspace{-0.25in}
\end{table}

We formulate the planning problem as a POMDP
(see \Cref{tab:BayesianSearchAndPOMDPs}) and solve the planning problem using a POMCPOW solver \cite{sunberg2018online}.
At each planing step, the robot selects the next
location to visit based on 
its current environment model. 
After moving to the next location, the robot
takes an image, collecting a set of (noisy) point measurements, which it feeds back into its GP model.
We assume a straight-line low-level motion planner for generating
trajectories from one location to the next.
We additionally enable a best-effort communication system between
robots and compare to other formulations (see \Cref{sssec:comms}). 
The planning process continues with each of the
$\nrobots$ robots until the allotted budgets have been reached, at which point
planning is complete and the robots can return to the base.
%
At that time, all measurements from all robots are compiled and used to produce the final quantile value estimates:
    $\estimatedquantilevaluesfinal = quantiles(\sensedvalues_\textrm{aggregate},\quantiles)$.

We now introduce several variations on the multirobot approach 
as illustrated in \Cref{fig:hero}.

\subsection{Initial Location Spread}
During informative path planning, robot teams will not cover the entire area
possible and are naturally affected by their starting area.  To study this, we
define an initial location spread parameter $\spread$ which varies from 0 (all robots
start at the same location) to 1 (robots are spread around the
entire workspace).  At spread 0.5, for example, the robots would start evenly
spaced in a rectangle half the size of the entire workspace. We choose initial locations for robots following a variation of Lloyd's algorithm \cite{lloyd1982least} which iteratively computes the centroids of an approximate Voronoi tessellation of the space and reassigns the locations to those centroids.
We perform the process for $100$ iterations using $100 \nrobots$ randomly sampled points. The result is $\nrobots$ locations approximately uniformly spread in the allotted workspace.

\subsection{Budget}
\label{sssec:budget}
To control for the fact that a team with more robots will in practice simply
have more planning steps, and for a fairer comparison to a single robot baseline,
we implement a variation on the budget constraint which we call
\textit{shared budget}: $\forall n \in \nrobots : B_n = B_T / \nrobots$.  
When $(B_T \hspace{-3px}
\mod{\nrobots}) \neq 0$, the remainder is split evenly among the most robots
possible.
The alternative is the default of a \textit{complete budget}, which gives each robot
the full budget, i.e. $\forall n \in \nrobots : B_n = B_T$.

\subsection{Communication}
\label{sssec:comms}
To investigate the impact of communication,
we distinguish between several versions of information sharing.
\textit{Full communication} assumes perfect, instantaneous communication,
which translates to every robot using the same shared GP model of the
environment, updated with every measurement by any robot,
or equivalently, centralized planning. 
%
At the other extreme is \textit{no communication}. In this case, each robot has its own environment model.
Essentially, this implementation is equivalent to
$\nrobots$ robots planning independently;
when new measurements are taken, they are only used to update that
robot's GP.
We believe a comparison to no communication provides a valuable baseline to understand what effect arises due solely to information sharing. 
In a practical sense, if separate research groups pool their robots, implementing coordinating mechanisms can demand significant time and effort. 
In that case, deploying the available robots with no behavioral changes may be simplest.

The third variation is \textit{stochastic communication}.
After taking measurements, a robot attempts to transmit the information (set of pixel locations and corresponding values) to
every other robot. We model each attempt as a Bernoulli trial and
assume data is successfully transmitted with sigmoidal probability $p_{\textrm{success}}$,
following \cite{clark2021queue} and where $distance$ is the distance between the
two robots, $\eta$ defines the sigmoid steepness,
and $r$ is the distance at which communication quality degrades past a threshold:
    $p_{\textrm{success}} = (1 + e^{\eta (distance - r)})^{-1}$.
We assume that if communication is successful between robots $n$ and $m$ at time
step $i$, all $\sensedlocations_i$ are received by $m$ (i.e., there are no partial or corrupted measurements transmitted).
The receiving robot's GP is updated with the received data.

We additionally compare to splitting the area into regions, or the \textit{partitioned} case.
This uses a Voronoi partition of the workspace based on the initial locations and restricts each robot to stay within its assigned region at all times.
No communication is enabled between partitioned robots.



\section{Experimental Setup}
We evaluate performance in simulation using two different real-world datasets. 
The datasets
are collected using a hyperspectral camera mounted on a drone flown over a freshwater lake. 
Each dataset is collected in the same area of the lake but on different days, leading to different algae distributions.
The robots are bounded in an area of approximately $80 \times 60$ meters.
We use the 400nm channel from the datasets as a proxy for algae and normalize the measured pixel intensity ($0-255$) to $[0, 1]$.
We test two sets of quantiles of interest in our experiments: quartiles $\quantiles = (0.25, 0.5, 0.75)$ and extrema $\quantiles = (0.9, 0.95, 0.99)$.
For each combination of parameters in all experiments, we run 2 seeds.
The available workspace is discretized into a 25x25 grid $\gtlocations$. The environment is unknown ahead of time to the robots.
We assume a communication model as described in \Cref{sssec:comms} with $\eta = 0.5$ and $r = 10$.
We additionally add zero-mean Gaussian noise with $\sigma_\textrm{noise} = 0.05$ to each measurement taken.
In the POMCPOW solver, we set number of rollouts per step to 100,
rollout depth to 4, and the planner discount factor to 0.8. 
The GP lengthscale is 12. The drone altitude is 7m and each image
is 25 pixels (measurements).

We select several parameters to vary in order to investigate their impact on
performance. 
These are listed in \Cref{tab:params}.
To measure performance, we report the root mean squared error (RMSE) between the ground
truth quantile values $\quantilevalues$ and the estimated quantile values
$\estimatedquantilevaluesfinal$ at the end of the surveys.

\begin{table}[t]
  \caption{Parameters studied in our experiments.}
  \begin{tabular}{lc} 
    \textit{Parameter} & \textit{Tested values} \\ 
    \hline 
    Initial location spread ($\spread$) & 0.0, 0.33, 0.66, 1.0 \\ 
    Total budget ($B_T$) & 10, 15, 30 \\
    Budget type  & \textit{complete, shared} \\
    Communication type &  \textit{none, stochastic, full, partitioned} \\ 
  \end{tabular}
  \label{tab:params}
  \vspace{-0.25in}
\end{table}

\subsection{Initial Location Spread}
\label{ssec:alpha-expt}
We first vary $\spread = \{0.0, 0.33, 0.66, 1.0\}$.
Each $\spread$ is tested on team sizes $\nrobots = \{2, 4, 8\}$, and 
we keep the budget constant at $B_T = 15$. 
We additionally compare each $\spread$ with no communication and with stochastic communication, but do not consider the single robot case, as $\spread$ and communication have no effect with only one robot.
To quantify the significance of the performance differences observed, we report results from the Wilcoxon signed-rank test \cite{wilcoxon1992individual}, which is a nonparametric version of the t-test and tests whether two paired samples originate from different distributions.

\subsection{Planning Budget}
\label{ssec:budget-expt}
Next, we vary $B_T = \{10, 15, 30\}$ in the complete budget case, setting $\spread$ to the constant value of $0.66$. 
We consider $\nrobots = 1$ as well as the multirobot teams.
In addition, we compare shared to complete budgets, setting $B_T = 15$.
In all cases, we enable stochastic communication between robots.

\subsection{Communication}
\label{ssec:comms-expt}
Finally, we compare performance with different $\nrobots$
across different levels of communication: full, stochastic, and none, and compare these to partitioning the space.
We hold previous parameters constant: $\spread = 0.66; B_T = 15$, and use $\nrobots = \{1, 2, 4, 8\}$.
Note that in the partitioned case, $\spread = 1.0$.


\section{Results}
We now present results from our experiments and include a discussion on their implications for real-world multirobot field work.
Boxplots show final RMSE between the estimated quantile values $\estimatedquantilevaluesfinal$ and the ground truth quantile values $\quantilevalues$ on the Y axis for an experiment, both aggregated across and refined by team sizes. RMSE is reported in terms of normalized pixel intensity.
Bars above indicate the Wilcoxon signed-rank significance levels for paired experiment groups.

\subsection{Initial Location Spread}
\label{ssec:alpha-results}
\begin{figure}[t!]
  \centering
 \includegraphics[width=\boxwidth,height=\boxheight]{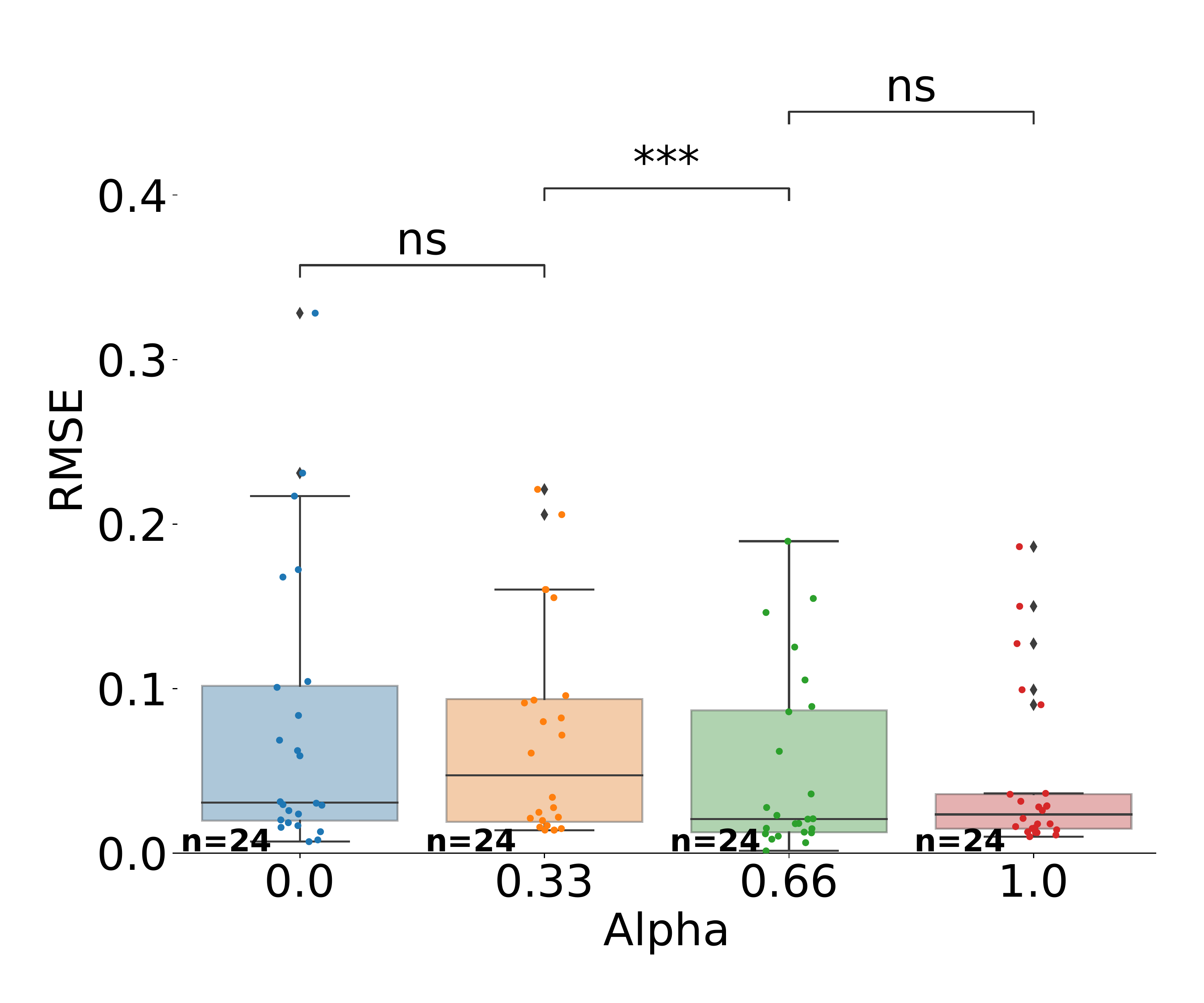}
   %
 \includegraphics[width=\boxwidth,height=\boxheight]{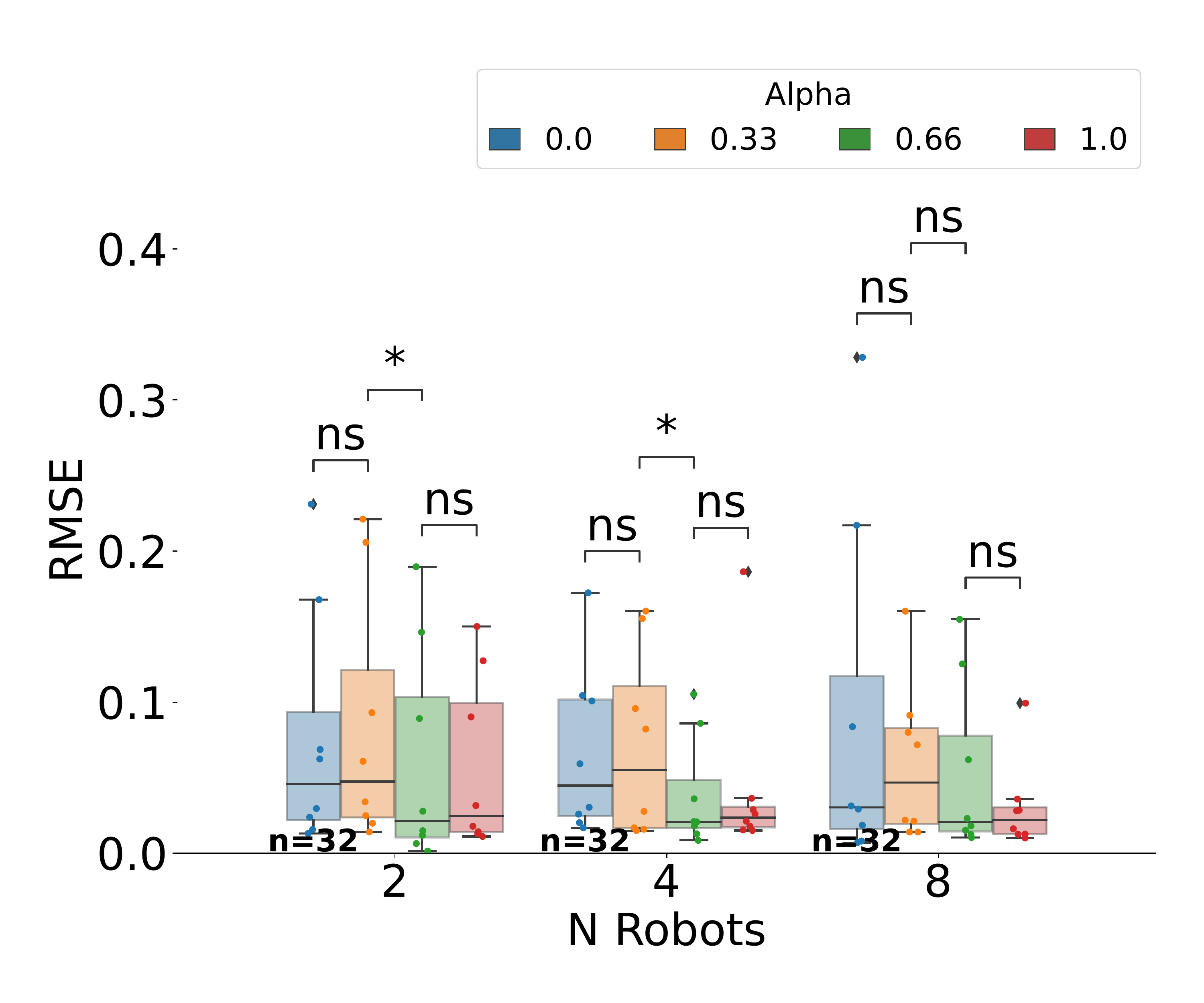}
  %
 \includegraphics[width=\boxwidth,height=\boxheight]{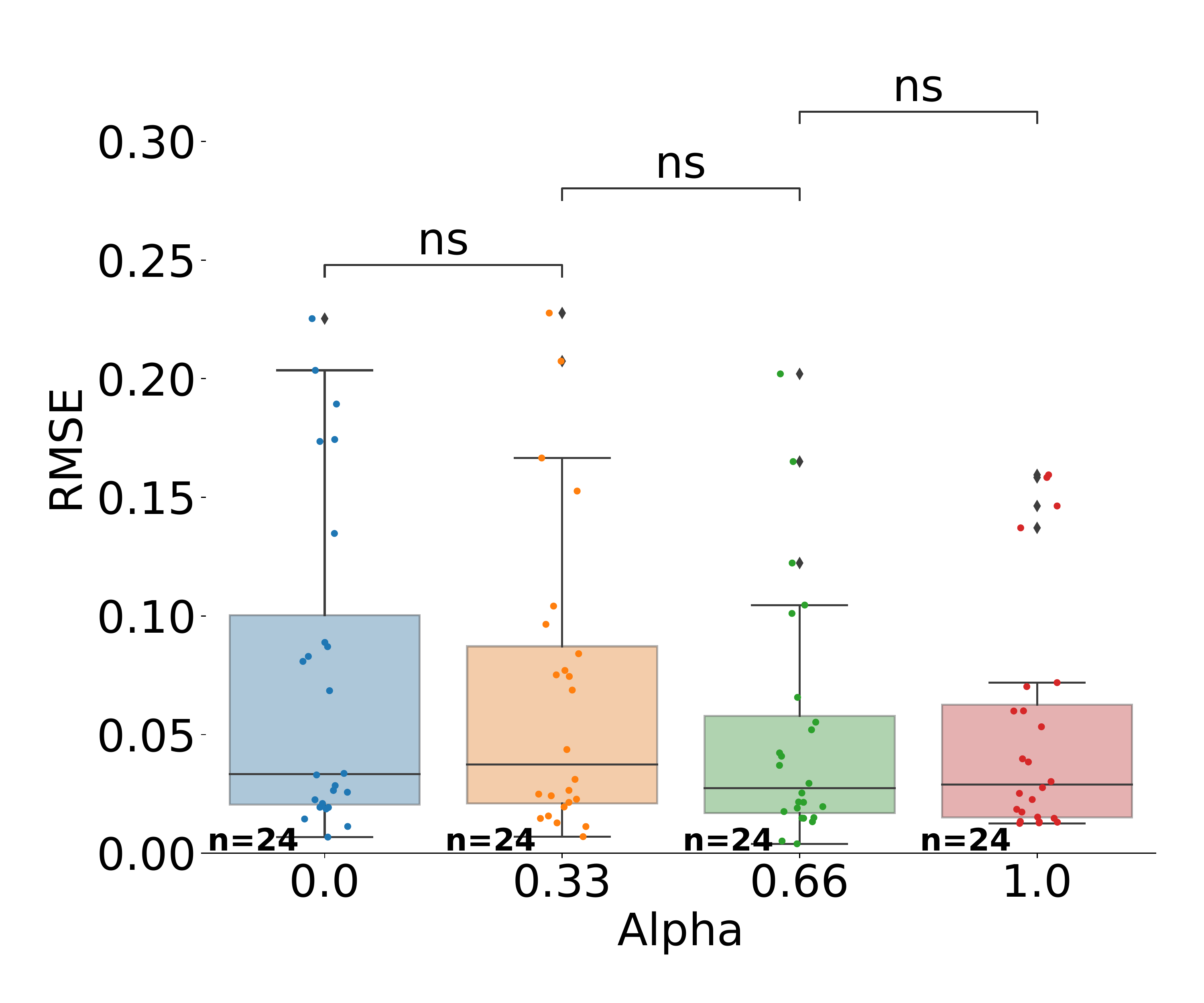}
  %
 \includegraphics[width=\boxwidth,height=\boxheight]{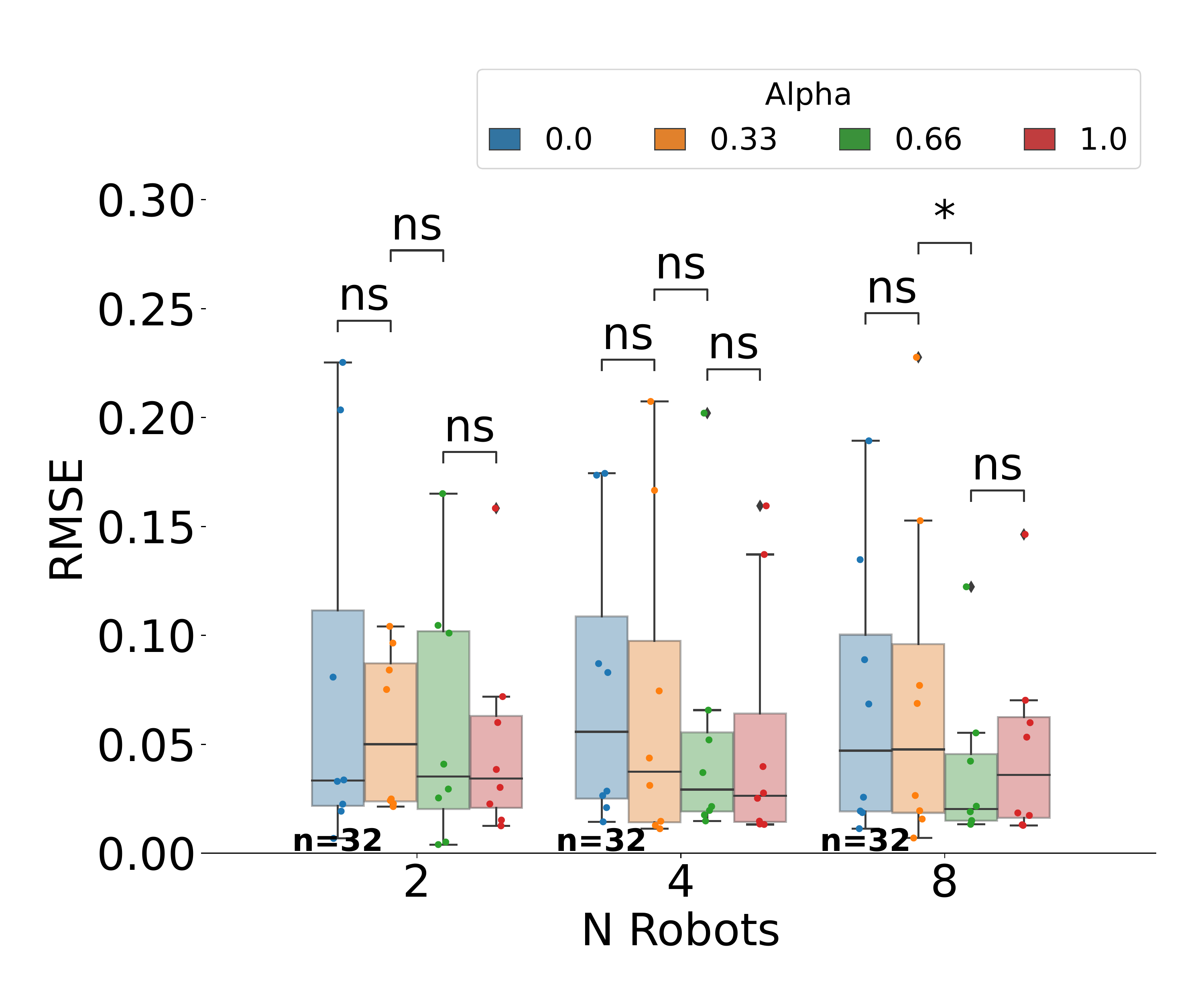}
  \caption{
  \textbf{Error vs. Initial location spread ($\spread$).}
  Top: No communication.
  Bottom: stochastic communication. 
  Right: Results further separated by team size. 
  Bars above indicate significance under the one-sided Wilcoxon signed-rank test; ***: $\pppsig$, *: $\psig$, ns: no significance
   \cite{florian_charlier_2022_7213391}.
}  
  \label{fig:alpha}
  \vspace{-0.25in}
  \looseness=-1
\end{figure}

\Cref{fig:alpha} shows the results of the experiments on $\spread$ using a one-sided Wilcoxon signed-rank test.
We observe that, in general, performance tends to improve (error decreases) with increasing $\spread$.
In particular, in the absence of stochastic communication as seen on the top row of \Cref{fig:alpha}, 
we see a statistically signficant performance improvement between $\spread = 0.33$ and $0.66$ ($\wpres{266}{0.001}$).
On the top right, further separated by $\nrobots$, we similarly observe 
a statistically significant improvement in performance between $\spread=0.33$ and $\spread=0.66$ for both $\nrobots=2$ and $\nrobots=4$ ($\wpres{33}{0.05}$, $\wpres{32}{0.05}$, respectively), 
but we do not see such an improvement between $\spread=0.66$ and $\spread=1.0$ in those cases.
In general, we also see that $\spread$ has a more drastic effect on performance the more robots there are, indicated by the decreasing maximum error. 
However, in general, there appear to be diminishing returns with bigger $\spread$.

The bottom row of \Cref{fig:alpha} shows results when robots communicate stochastically with each other at different initial location spreads.
Although we see the same general trend on left, in this case, we do not see a statistically significant drop in error. 
When controlled for team size on the right, we again see increasing $\spread$ resulting in improved performance 
as indicated by lower median and maximum errors.
Compared to the case where there is no communication, we do not notice the same dependency on $\nrobots$ for the effect of $\spread$; in other words, the decrease in error due to $\spread$ appears relatively similar across different values of $\nrobots$ (shown by relative median and maximum errors).
We additionally note that for $\nrobots=8$, there is a statistically significant drop in error from $\spread = 0.33$ to $0.66$ ($\wpres{33}{0.05}$).
Based on these results, 
if robots must be deployed near each other but cannot communicate, then a small robot team may do just as well as a larger team, and vice versa; a small team that cannot communicate does not need to be spread very far to be effective.
On the other hand, a larger spread combined with a large team size will be beneficial in the absence of inter-robot communication, but there will be marginal returns with larger spread.
If communication is enabled, a larger spread will generally always be beneficial (although again with marginal returns), regardless of the team size.

\subsection{Planning Budget}

\begin{figure}
  \centering
 \includegraphics[width=\boxwidth,height=\boxheight]{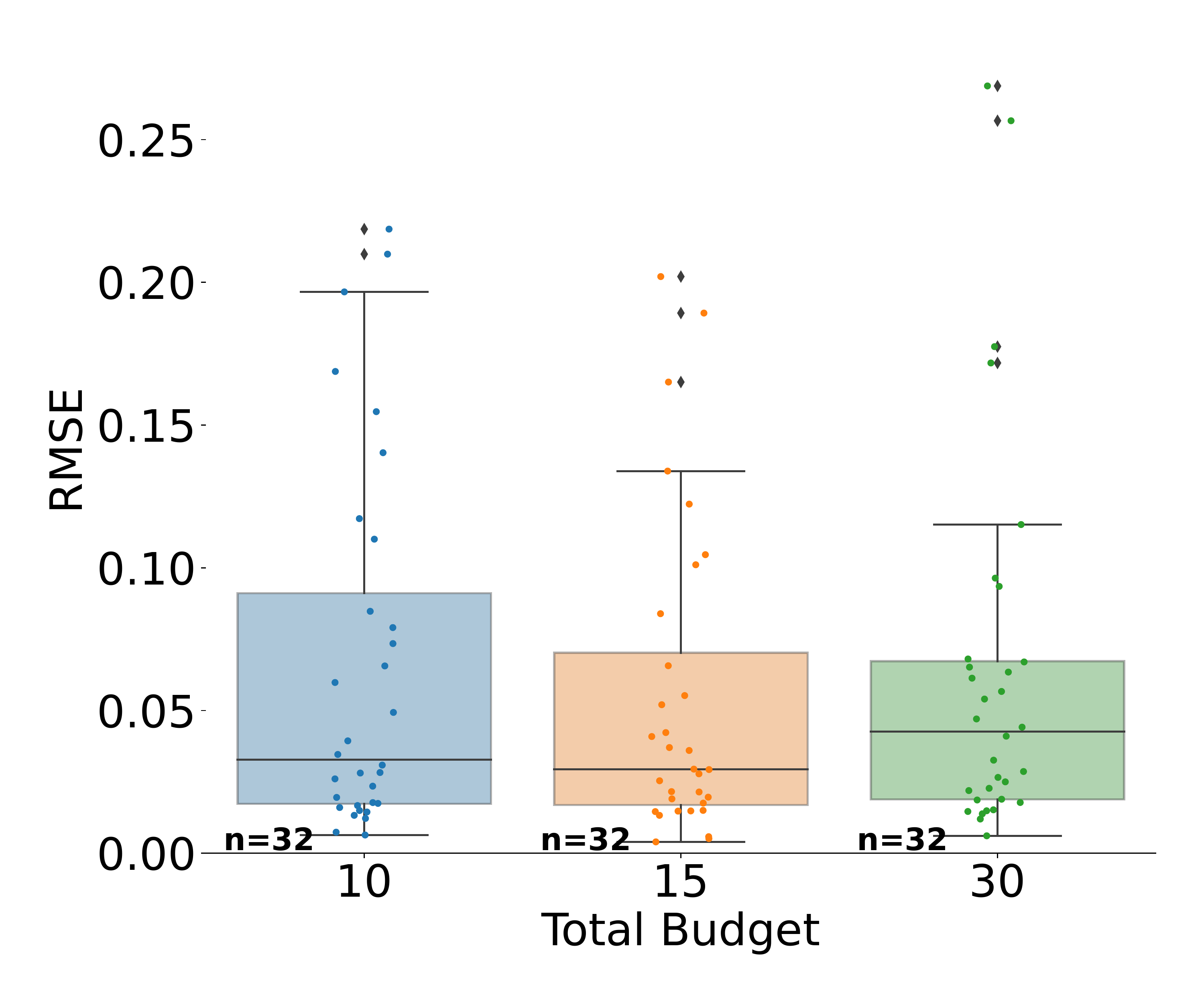}
  %
 \includegraphics[width=\boxwidth,height=\boxheight]{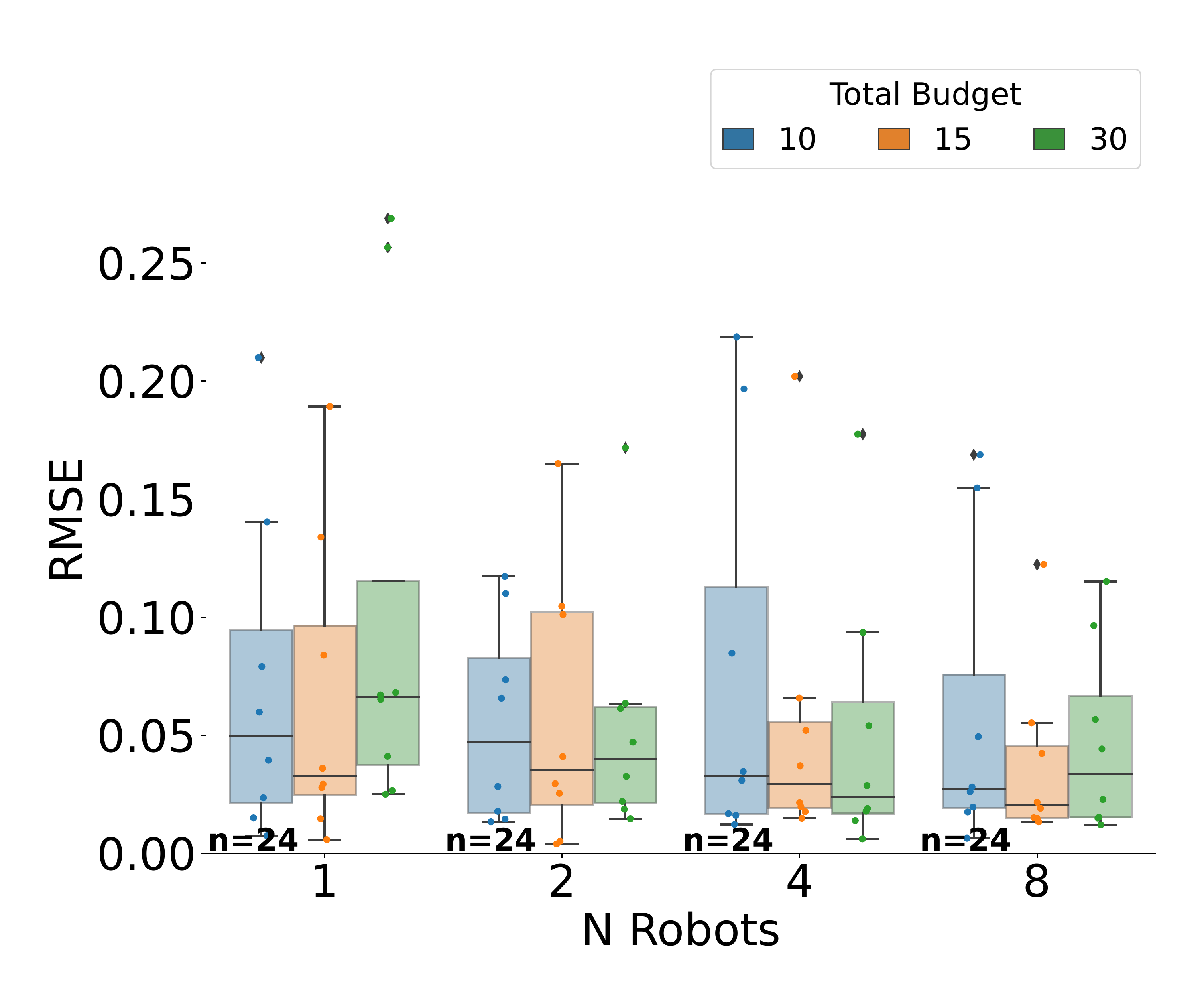}
  \caption{
  \textbf{Error vs. Budget (number of planning steps $B_T$).} 
  Right shows same results further separated by team size. 
  }
  \label{fig:budget}
  \vspace{-0.15in}
\end{figure}

\begin{figure}
  \centering
 \includegraphics[width=\boxwidth,height=\boxheight]{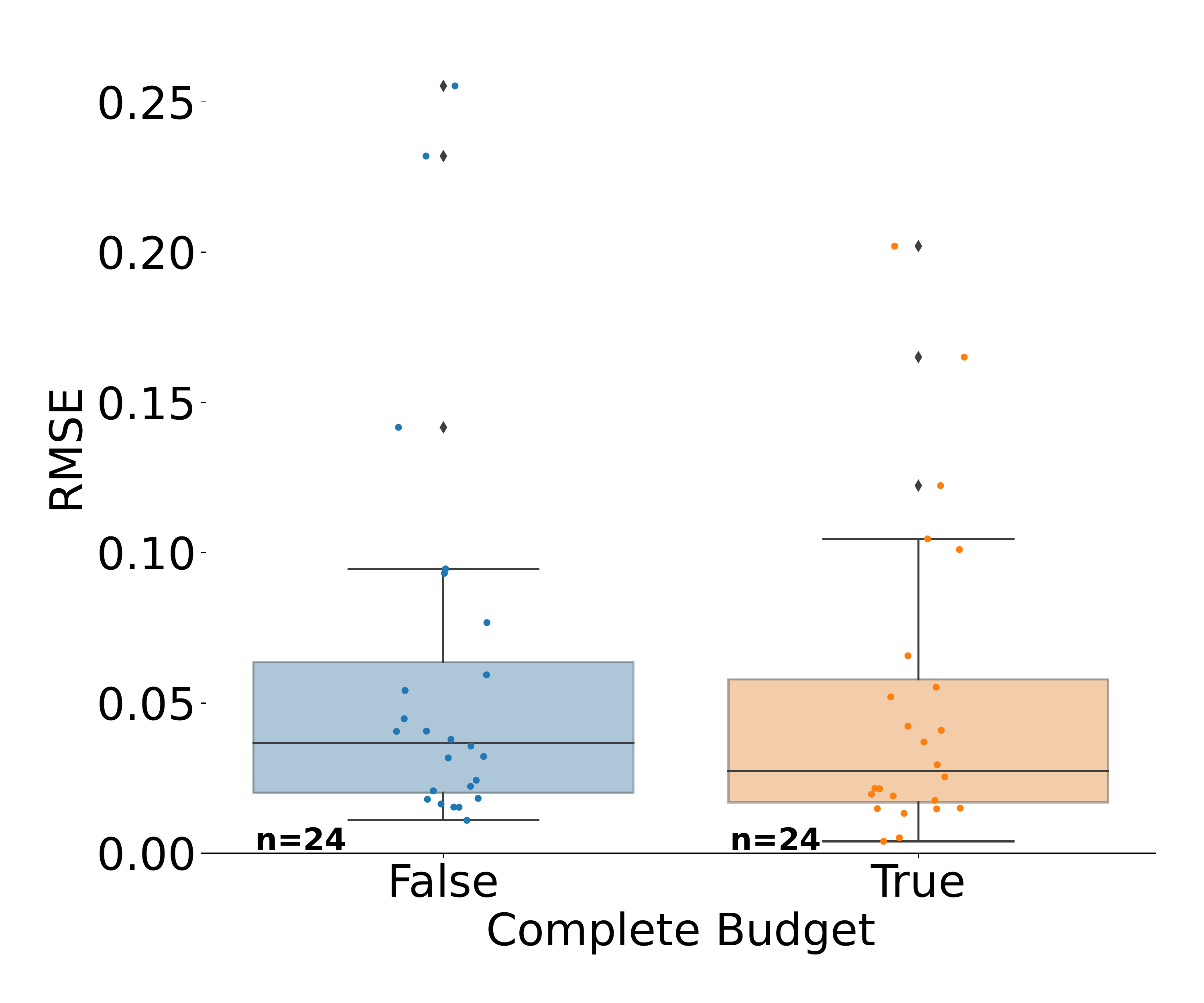}
  %
 \includegraphics[width=\boxwidth,height=\boxheight]{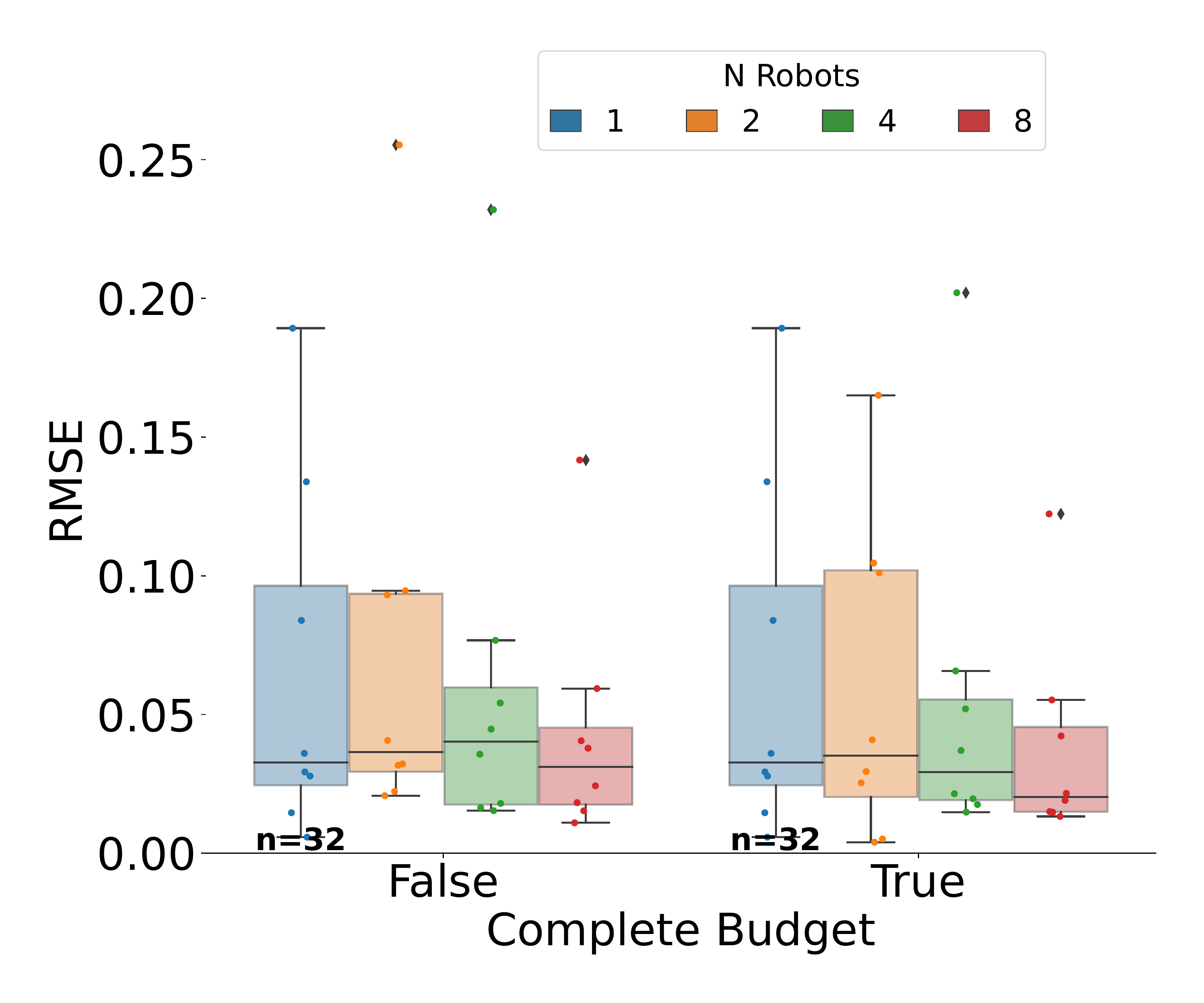}
  \caption{\textbf{Error vs. Budget Type (Shared and Complete).}
  Here, $B_T=15$.
  Left: Shared budget means the combined budget is $B_T$ for any $\nrobots$, while complete budget means it is $\nrobots \cdot B_T$. Note $\nrobots=1$ data not included here since budget type has no effect.
  Right: Same results, further separated by team size. 
  $\nrobots=1$ (blue) shown for comparison.
  \vspace{-15px}
  }
  \label{fig:complete-budget}
\end{figure}

Observing the maximum errors on the left of \Cref{fig:budget} suggests that in general,
a larger  $B_T$ leads to better performance, which is expected. 
What we observe on the right, however, is that the improvement actually comes from an increase in $\nrobots$ as opposed to in $B_T$, which is somewhat surprising (indicated by lower median error for higher $\nrobots$).

To investigate this further, we control for total path length $\sum_n B_n$ and show results for shared and complete budgets in
\Cref{fig:complete-budget}.
With $B_T = 15$ and $\spread = 0.66$ in this experiment, we see similar performance between shared and complete budgets, which is again counterintuitive, as we would generally expect improvement with complete budgets.
However, we uncover the real effect when we further control for $\nrobots$, on the right. 
With a complete budget, more robots (i.e., more steps) leads to better performance as well, as expected. 
What is notable is that, although we see similar median errors for all cases, the improvement in maximum error is also observed when the total path length is held constant.
This is particularly remarkable because in the $\nrobots = 8$ case, this equates to each robot taking at most 2 steps.
Given these results, it appears that having more robots, even with restricted budgets (e.g., time, path length), is preferable to having one robot with a larger budget.


\subsection{Communication}

\begin{figure}
  \centering
  \includegraphics[width=\boxwidth,height=\boxheight]{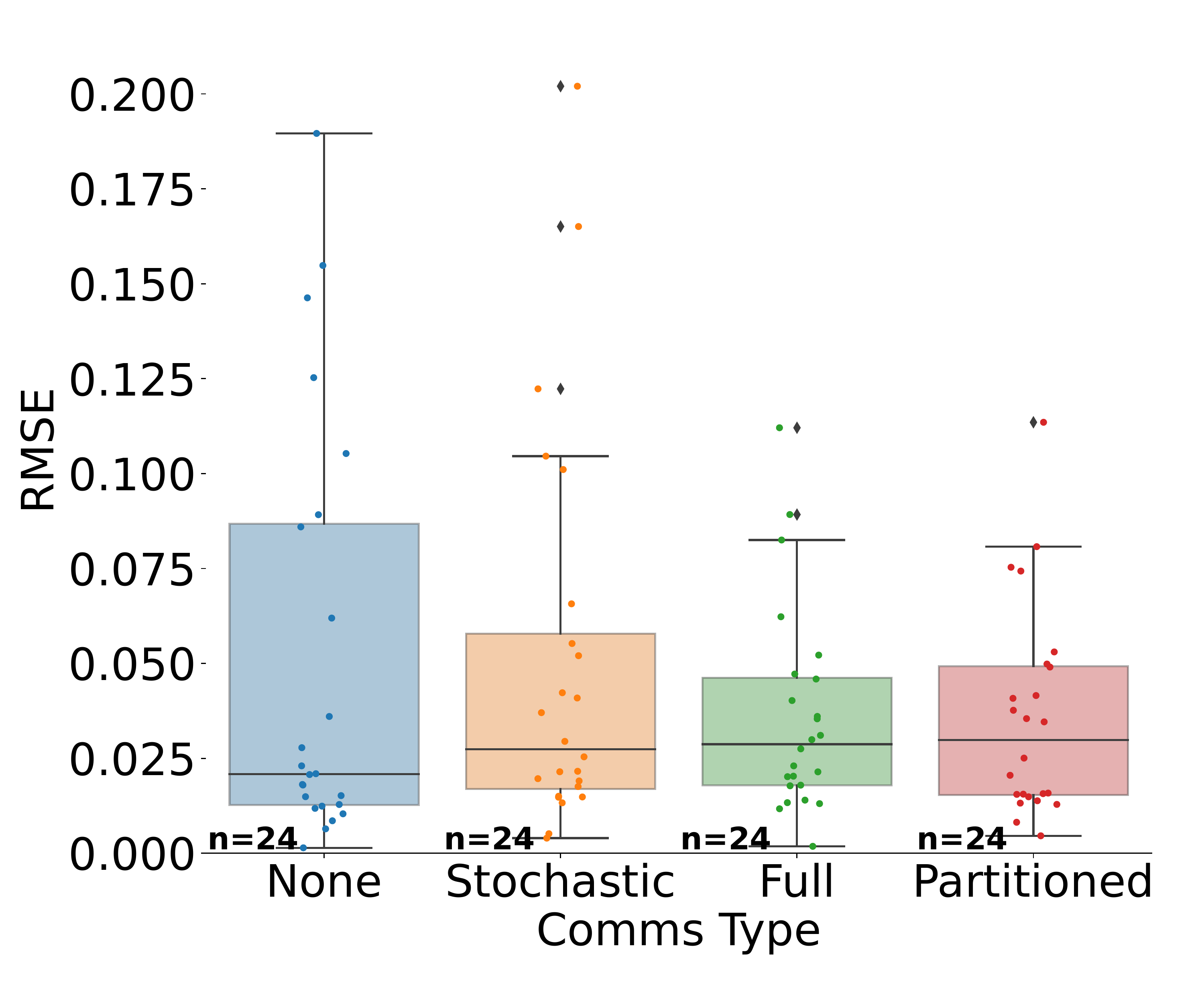}
  \includegraphics[width=\boxwidth,height=\boxheight]{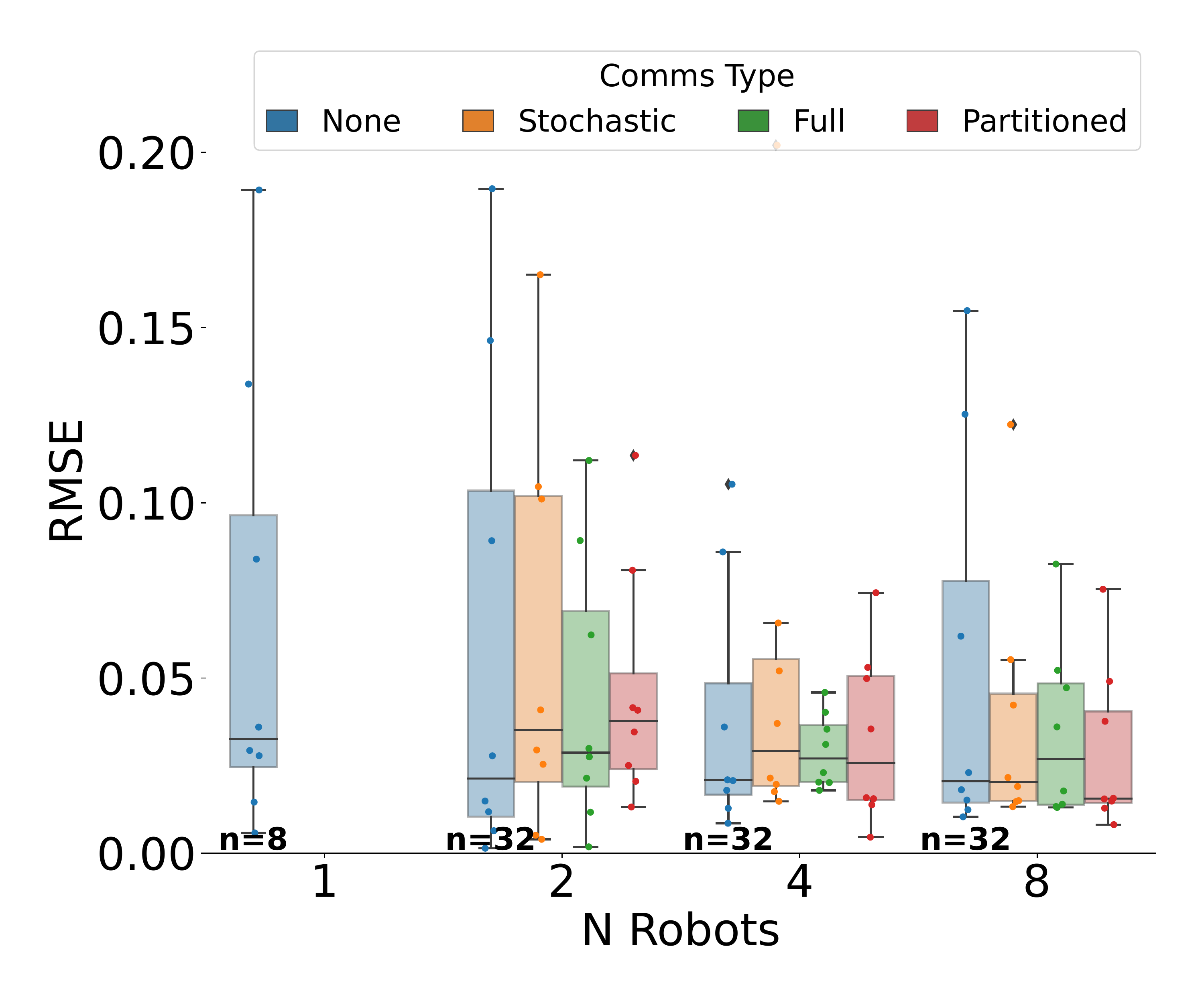} 
  \caption{
  \textbf{Error vs. Communication type.}
  Left: Aggregated across different $\nrobots$.
  Note, results for $\nrobots = 1$ not included, as they are not affected by the variations.
  Right: Results further separated by team size. $\nrobots=1$ included for comparison.
  }
  \label{fig:comms}
  \vspace{-0.17in}
\end{figure}

\Cref{fig:comms} shows the quantile estimation error when the robot team uses different communication styles. 
Note that on the right, we include the single robot case as a baseline.
On the left, we see that overall, 
all communication types have a median performance around the same as no communication.
We see that more communication does, however, lead to improved performance, indicated by smaller upper quartile errors. 
This is likely due to the fact the robots can avoid multiple measurements of locations that are not of interest.

On the right, we see that stochastic communication does not have much of an effect compared to no communication when $\nrobots= 2$ or $4$, as seen in the similar median and upper quartile errors.
At $\nrobots = 8$, we see that 
both the upper quartile error and the maximum error for no communication is much higher. 
This suggests that enabling communication is more beneficial for larger teams.
We also see that in general, partitioning achieves similar median error as when there is stochastic or full communication, as well as maximum error particularly for larger teams. 
Given that here, the partitioned case used $\spread = 1.0$ rather than $0.66$, this suggests that partitioning the space is an effective strategy if the entire area is accessible for deploying robots but communication is not possible.
Based on these results, performance can be improved by enabling inter-robot communication, even when imperfect, and it is particularly beneficial with large teams.

\begin{figure}
  \centering
  \includegraphics[trim={0.5cm 3cm 0.5cm 3.5cm}, clip, width=\boxwidth]{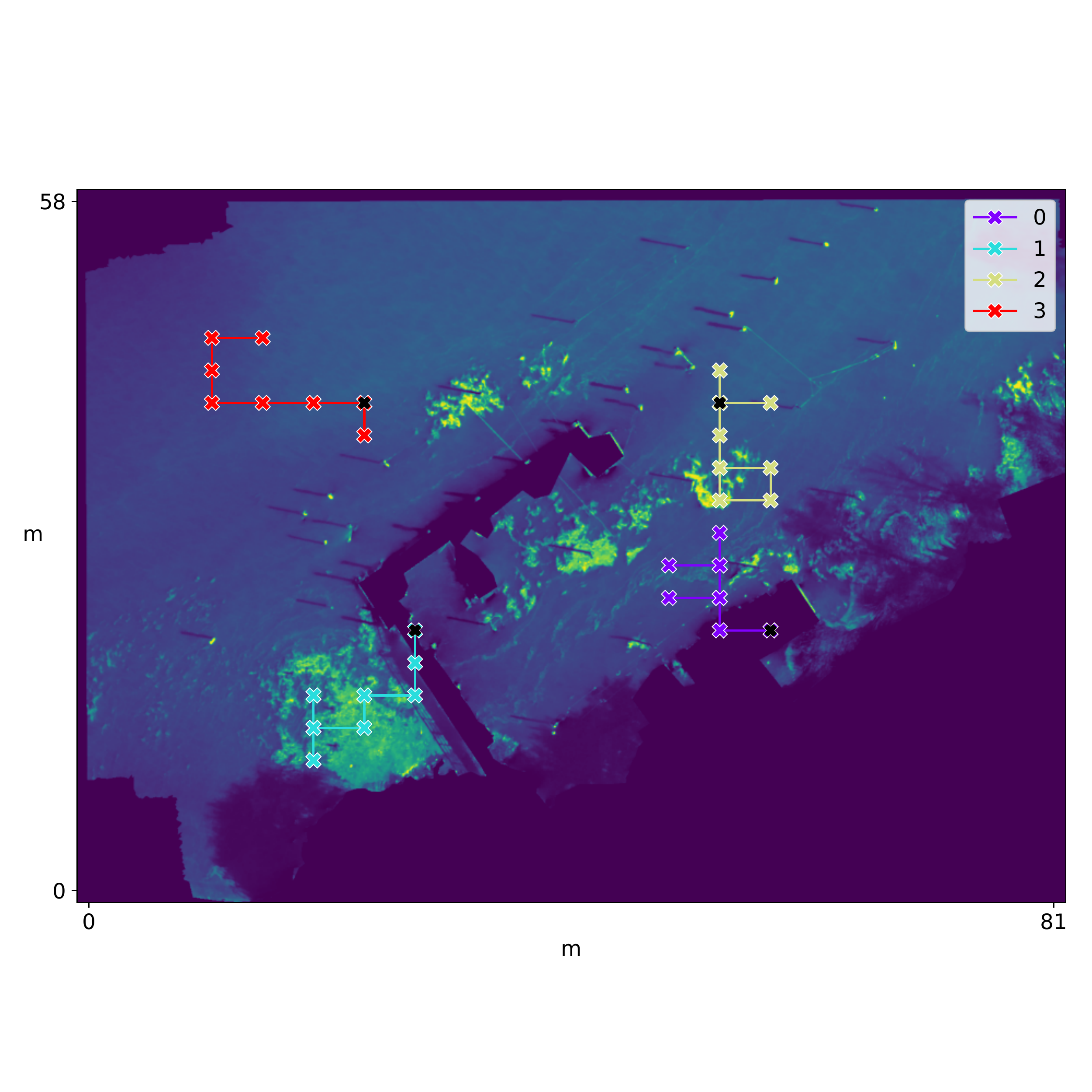}
  \includegraphics[trim={0.5cm 3cm 0.5cm 3.5cm}, clip, width=\boxwidth]{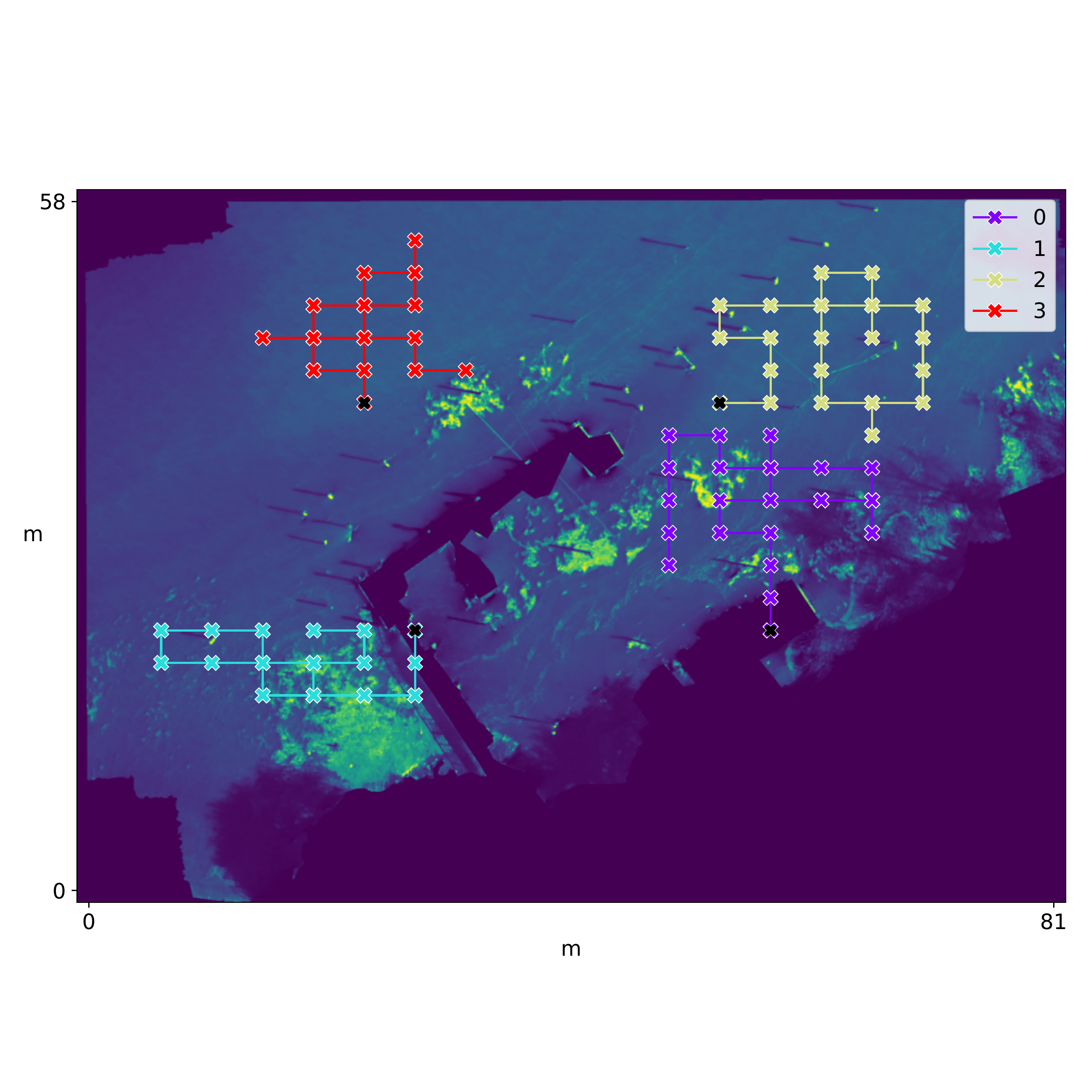}
  \caption{
  Example paths of a four-robot team with $B_T=10$ (left) and $B_T=30$ (right).
  }
  \label{fig:raster-increase-budget}
  \vspace{-0.25in}
\end{figure}

\begin{figure}
  \centering
  \includegraphics[trim={0.5cm 3cm 0.5cm 3.5cm}, clip, width=\boxwidth]{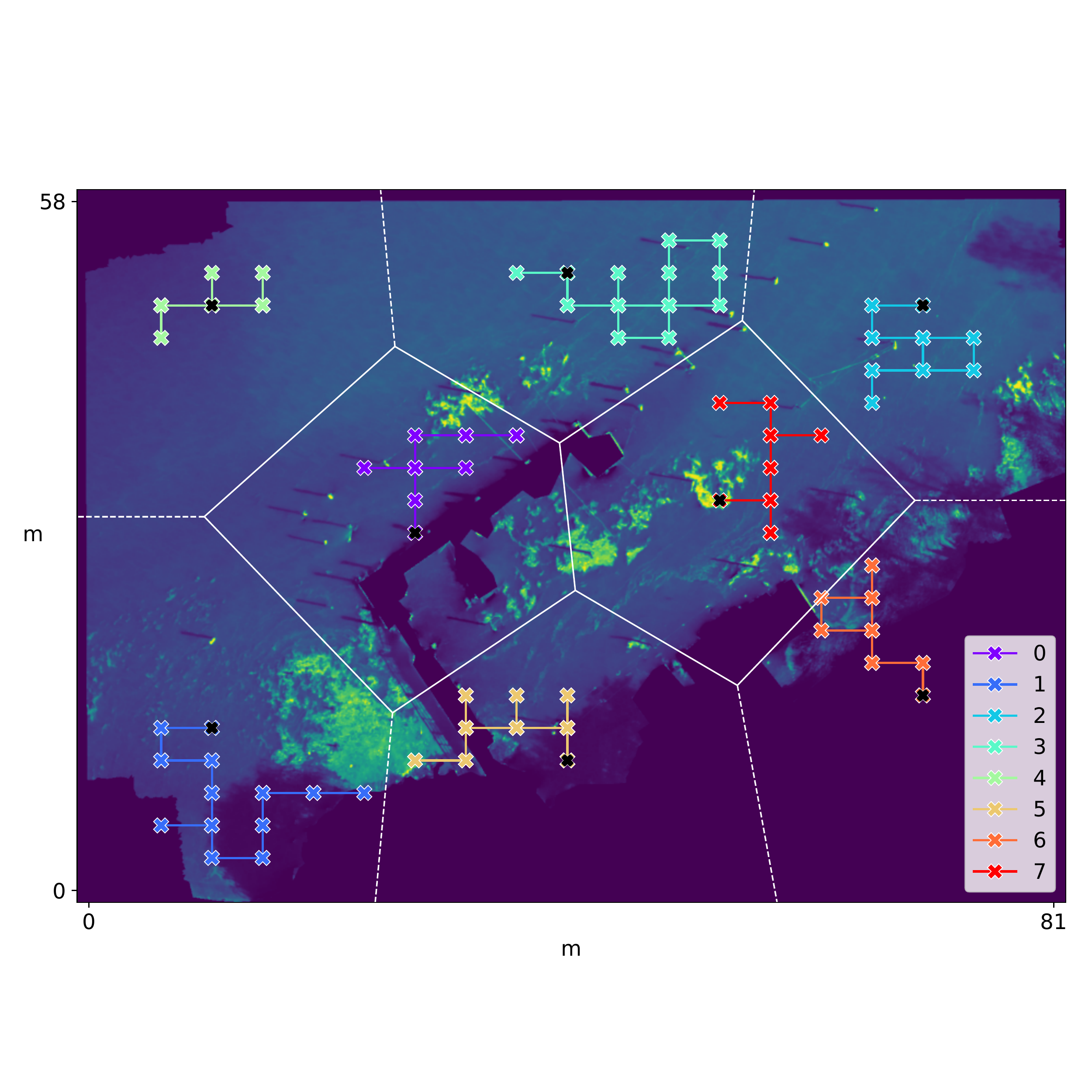}
  \includegraphics[trim={0.5cm 3cm 0.5cm 3.5cm}, clip, width=\boxwidth]{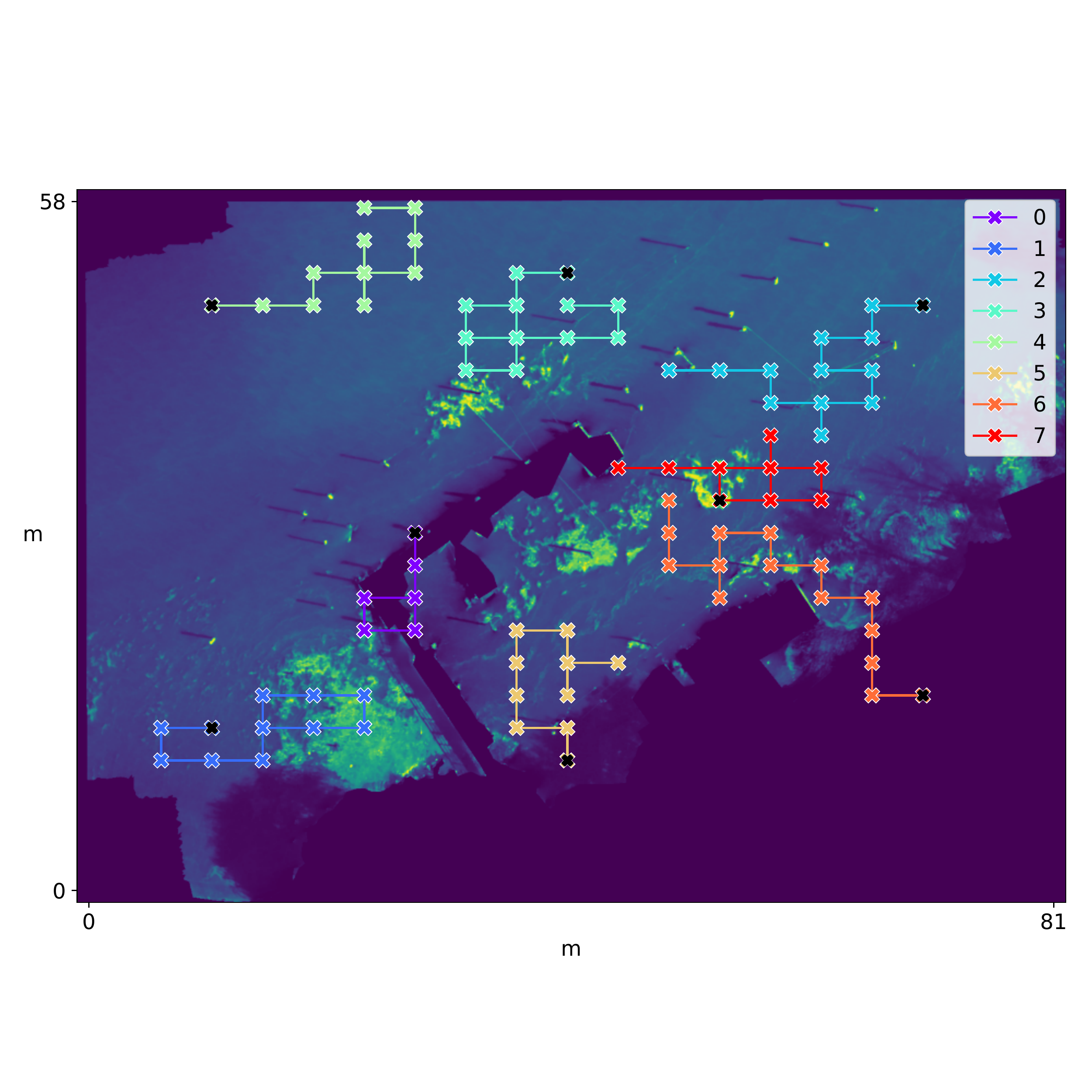}
  \caption{
  Example paths of an 8-robot team under partitions (left) and no communication (right).
  }
  \label{fig:raster-partition}
  \vspace{-0.25in}
\end{figure}

Finally, we show some example plans produced during these experiments as a visualization of the task in \Cref{fig:raster-increase-budget,,fig:raster-partition}.
The robots are shown in different colors and the paths they produce are shown as connected crosses.
The initial locations are marked in black. If the robots were partitioned, the assigned regions are designated with white lines.
The background of the figures is the orthomosaic of the hyperspectral images
previously collected by drones in real aquatic environments.

\section{Discussion}
We now summarize the main conclusions drawn from the presented results, and how they may influence and impact different field robotics setups and future work.

With regards to the initial location spread, a large spread appears to be beneficial when $\nrobots$ is large, but not when $\nrobots$ is small.
Additionally, 
a larger spread gives marginal returns.
This may be particularly valuable information when planning real field deployments, where time and monetary costs to getting many robots to spatially diverse areas exist; avoiding unnecessary costs is crucial.
In addition, if $\spread$ is restricted to very low values by the survey resources or setup, then using more robots may help regain some performance.

Notably, in terms of budget $B_T$, a larger budget in and of itself does not significantly improve performance for a fixed $\nrobots$ (assuming a moderate initial spread). What we observe instead is a stronger dependency on $\nrobots$ itself; therefore, a larger number of robots, each with a restricted budget, will likely produce better quantile estimates than a single robot with a $B_T$ equal to the total budget of all of them.

Our results also indicate that enabling communication is generally beneficial to the quantile estimation problem, and in particular for larger groups. 
We note that partitioning the space into disjoint areas and restricting each robot to one area can be an effective alternative strategy,
particularly if inter-robot communication is not an option.

Looking forward,
it would be interesting to consider communication protocols that take into account the potential utility of a message, or with more nuanced communication models based on particular hardware capabilities.
We are also interested in how this scales to larger team sizes.
In the context of lake monitoring and other environmental monitoring tasks, swarms are not typically deployed;
however, for large-scale studies, it may be useful to understand that impact.

In practice, scientist teams will need to weigh the benefit of lower error with the practicality of acquiring, coordinating, and deploying more robots for their specific application.



\section{Conclusion}
In this work, we have presented the first study on multirobot quantile estimation
for environmental analysis. We investigate the impact of multiple robots on
quantile estimation accuracy, as well as the effect of initial location spread, planning budget, and communication.
We measure how well different combinations of parameters perform in terms of error in quantile value estimates, 
and we further provide statistical results quantifying the significance of the differences observed.
This work is an important first step toward characterizing the elements of an effective
multirobot system for environmental analysis and has potential to help
scientists interested in larger or collaborative survey projects to better
understand the benefits and drawbacks of different experiment design choices. In
turn, this may improve environmental monitoring and conservation
efforts.

\addtolength{\textheight}{-1cm}   



%
%


\bibliographystyle{IEEEtran}
\bibliography{IEEEabrv,references}

\end{document}